\documentclass[conference]{IEEEtran}
\IEEEoverridecommandlockouts
\usepackage{cite}
\usepackage{amsmath,amssymb,amsfonts}
\usepackage{algorithmic}
\usepackage{graphicx}
\usepackage{textcomp}
\usepackage{xcolor}
\usepackage{multirow}
\usepackage{hyperref}
\def\BibTeX{{\rm B\kern-.05em{\sc i\kern-.025em b}\kern-.08em
    T\kern-.1667em\lower.7ex\hbox{E}\kern-.125emX}}
\begin{document}

\title{Online Adaptation for Enhancing Imitation Learning Policies}

\author{\IEEEauthorblockN{1\textsuperscript{st} Federico Malato}
\IEEEauthorblockA{\textit{School of Computing} \\
\textit{University of Eastern Finland}\\
Joensuu, Finland \\
federico.malato@uef.fi}
\and
\IEEEauthorblockN{2\textsuperscript{nd} Ville Hautam\"aki}
\IEEEauthorblockA{\textit{School of Computing} \\
\textit{University of Eastern Finland}\\
Joensuu, Finland \\
ville.hautamaki@uef.fi}
}

\maketitle

\begin{abstract}
Imitation learning enables autonomous agents to learn from human examples, without the need for a reward signal. Still, if the provided dataset does not encapsulate the task correctly, or when the task is too complex to be modeled, such agents fail to reproduce the expert policy. We propose to recover from these failures through online adaptation. Our approach combines the action proposal coming from a pre-trained policy with relevant experience recorded by an expert. The combination results in an adapted action that closely follows the expert. Our experiments show that an adapted agent performs better than its pure imitation learning counterpart. Notably, adapted agents can achieve reasonable performance even when the base, non-adapted policy catastrophically fails.
\end{abstract}


\section{Introduction}
\emph{Reinforcement learning} (RL)~\cite{suttonbarto} and \emph{Deep reinforcement learning} (DRL)~\cite{drl} have recently gained momentum as a consequence of notable breakthroughs in policy learning~\cite{vpt, dreamer2, dreamer3}, and following the introduction of \emph{reinforcement learning from human feedback} (RLHF)~\cite{rlhf} for fine-tuning large language models (LLMs). In previous research, RL and DRL have been successfully applied in several domains, ranging from (and not limited to) playing video games~\cite{kanervisto2020a, kanervisto2020b, starcraft}, autonomous driving~\cite{driving}, to physics~\cite{tomahawk}.

Despite these incredible results, several challenges remain open~\cite{drlsurvey}. Among those, specifying a reward signal for complex, structured tasks is one of the most prominent. A common approach to address this problem is \emph{imitation learning} (IL)~\cite{il}, that is, agents learn to act from an expert demonstrating the task, without a reward signal. Perhaps, the most known example of IL algorithm is \emph{behavioral cloning} (BC)~\cite{bc}. In BC, a policy is learned via supervised learning on the observation-action pairs of the expert dataset. Another notable IL-derived framework is \emph{inverse reinforcement learning} (IRL)~\cite{irl, ng2000}, where a policy is trained using standard RL on a reward model inferred from the expert trajectories.

\begin{figure}[htbp]
\centerline{\includegraphics[width=\columnwidth]{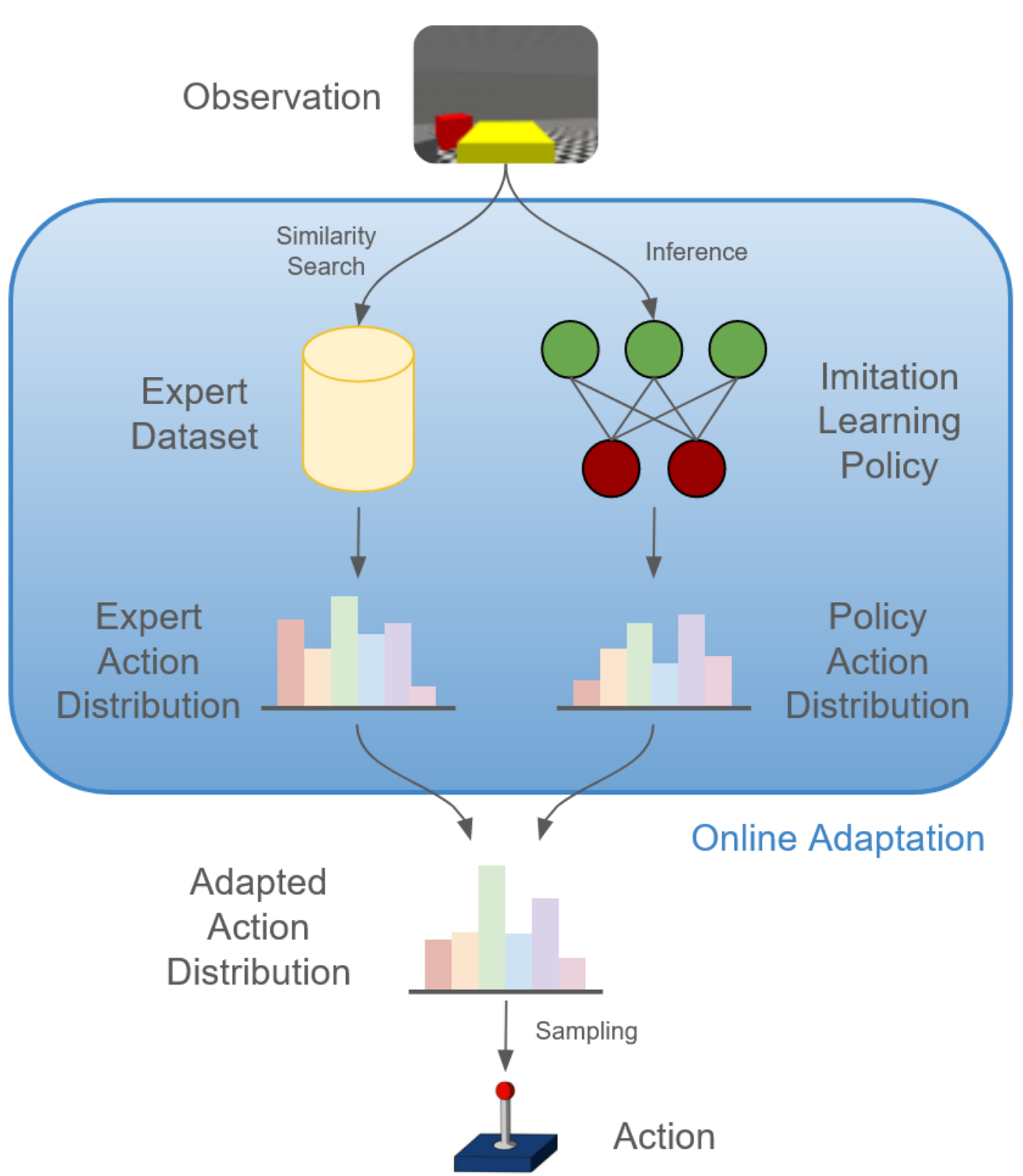}}
\caption{A visual explanation of our proposed method. At timestep $t$, the current observation is fed to the imitation learning policy network to obtain a policy action distribution. Concurrently, we retrieve a number of resembling frames from the expert data and compute an expert action distribution. The two distributions are then combined to obtain a joint, adapted action distribution. Finally, the action is selected by sampling from the joint distribution.}
\label{fig:abstract}
\end{figure}

While IL policies extend RL agents to scenarios where no reward function can be designed, they also suffer from a number of drawbacks. For instance, BC suffers distributional shift and causal confusion~\cite{bc}, while IRL is never guaranteed to learn an optimal reward model~\cite{irl}. A solution to these problems is \emph{adversarial imitation learning}, where a policy is trained by learning two networks simultaneously in an adversarial fashion. The most notable examples of adversarial algorithms are {\em generative adversarial imitation learning} (GAIL)~\cite{gail} and {\em adversarial inverse reinforcement learning} (AIRL)~\cite{airl}. GAIL and AIRL take inspiration from generative adversarial networks (GANs)~\cite{gan}. Specifically, both algorithms train a policy along with a discriminator that tries to distinguish between policy-generated and expert-generated trajectories~\cite{gail, airl}. AIRL differs from GAIL as it also derives a reward function. Both algorithms have significantly improved the performance of autonomous agents in a wide range of tasks. Still, adversarial training is unstable and sample inefficient~\cite{gan, gail, airl}, that is, an adversarial agent might never learn a useful policy, and will typically require very long training procedures.

Additionally, a range of challenges still require a general solution. For instance, autonomous agents should be able to model long-term, causal relationships and plan their actions accordingly~\cite{suttonbarto, planning}. Model-based RL attempts to solve these problems by learning an implicit representation of the environment~\cite{planning}. More recently, OpenAI proposed Video PreTraining (VPT)~\cite{vpt}, a transformer-based~\cite{transformers}, causal model to play Minecraft from human demonstrations. VPT was capable of solving long-standing challenges such as the MineRL Diamond challenge~\cite{minerl}. Similarly, recently proposed Dreamer architectures~\cite{dreamer2, dreamer3} merge several network architectures and combine IL and DRL methods to learn a \emph{world model} to operate informed decisions. 

While agents such as VPT and Dreamer indeed achieve outstanding results, they are bounded to use complex architectures~\cite{vpt, dreamer2, dreamer3} or need massive datasets to be trained successfully~\cite{vpt}. Therefore, their usability in real world scenarios remain uncertain.

In real world use cases, usually gathering data is expensive and demands massive resources~\cite{minedojo, bedd, genai}. Hence, autonomous agents should be able to learn from small dataset, while being robust to unpredictable conditions and aligned to human needs. Previous research in this direction have leveraged search as a way to reliably and efficiently select actions~\cite{pari2021, zip}. For example, a robotic arm in a low-dimensional domain with continuous action space can average over a set of retrieved relevant actions~\cite{pari2021}. In the case of an open-world, high-dimensional visual domain with discrete actions, copying a sequence of actions from relevant past experience of an expert has been proven successful~\cite{zip}. Despite showing robust performance, such methods lack real time adaptability to unpredictable conditions.

Inspired by this previous research, we propose {\em Bayesian online adaptation} (BOA), an efficient technique to improve an IL agent action selection process using search, requiring little to none scaling in network complexity. Our approach is explained visually in Fig.~\ref{fig:abstract}. BOA leverages Bayesian statistics and search to improve the performance of pure IL agents. Additionally, our method allows a partial explanation of the action selection process, hence improving interpretability of the model.

\section{Preliminaries}
\label{sec:prelim}
\paragraph{Reinforcement Learning}
We model our control problem as a {\em partially observable Markov decision problem} (POMDP) as a $7$-tuple $(S, A, T, R, \Omega, O, \gamma)$ where $S~\subset~\mathbb{R}^d$ is the state space, $A$ is the action space, $T: S \times A \rightarrow S$ is the transition dynamics, $R:S \times A \rightarrow \mathbb{R}$ is the reward function, $\Omega$ is a set of observations, $O$ is a set of conditional observation probabilities and $\gamma \in [0, 1)$ is the discount factor.

\paragraph{Imitation Learning}
In the IL scenario, the reward function $R$ and the transition dynamics $T$ are unknown. As such, a policy can not acquire relevant experience by interacting with the environment. Instead, a dataset of observation-action pairs $\mathcal{D} = \{(o_t, a_t)\}, \mathcal{D} \subset S \times A$ with $t \in [0, \tau], \tau \in \mathbb{N}$ is provided by an expert demonstrating the task. The general aim of an IL policy is to minimize a loss $\mathcal{L}: A \times A \rightarrow \mathbb{R}$ that describes the difference between the policy predicted actions and the expert actions.

\paragraph{Multinomial Distribution}
The Multinomial distribution is a parametric, discrete distribution characterized by two parameters $K$ and $N$. $K$ is often referred to as \emph{classes} or \emph{categories}, while $N$ indicates the number of trials. Given a random variable $X\sim(N; p_1, \ldots, p_K)$, the distribution has a discrete probability density function of the form
\begin{equation}
\mathbb{P}(X=x) = \frac{N!}{\prod_{i=1}^{K}x_{i}!}\prod_{i=1}^{K}p_{i}^{x_i}
\end{equation}
Whenever $N=1$, the Multinomial distribution becomes a \emph{Categorical} distribution.
\paragraph{Dirichlet Distribution}
The Dirichlet distribution is a parametric distribution defined by a scalar parameter $K$ and a vector parameter $\bold{\alpha}$, called \emph{categories} and \emph{concentration} respectively. The probability density function is
\begin{equation}
f(x;\bold{\alpha}) = \frac{1}{B(\bold{\alpha})}\prod_{i=1}^{K}{x_{i}^{\alpha_i - 1}}
\end{equation}
where $B(\bold{\alpha})$ is the multivariate beta function.

Among its other properties, the Dirichlet distribution is the \emph{conjugate prior} of the Multinomial distribution. That is, if in Bayesian inference the prior follows a Dirichlet distribution and the likelihood follows Multinomial distribution, then the posterior is known to also follow a Dirichlet distribution.

\section{Proposed Method}
\label{sec:methods}
Our method leverages Bayesian inference to update the beliefs of an autonomous agent in real-time. The intuition is as follows: in general, an IL agent tries to imitate the expert's action distribution, given an observation. It follows that IL works as long as the dataset fully encapsulates the dynamics of a task. Whenever this condition is not satisfied, an IL agent is bound to either fail or show sub-optimal behavior~\cite{bc, il}. To mitigate this problem, we provide a learning-based agent with a minibatch of expert solutions for a particular state. Then, we infer the probable action of the expert and update the agent action distribution accordingly.

\begin{figure*}[htbp]
\centerline{\includegraphics[width=\textwidth]{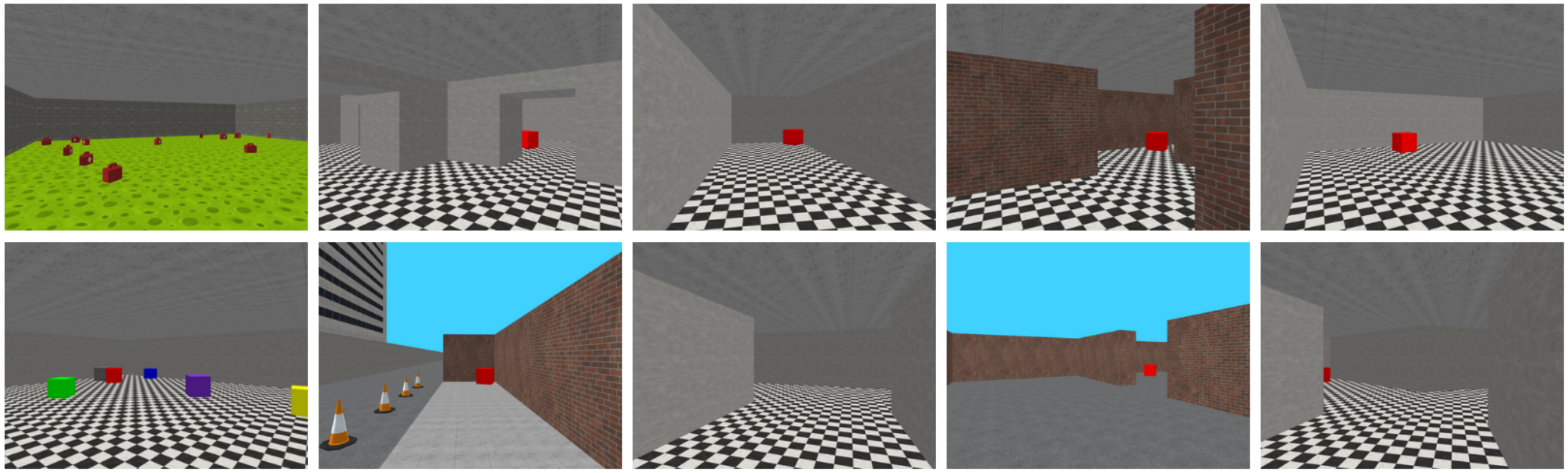}}
\caption{Screenshots from the $10$ MiniWorld environments used in our experiments. MiniWorld uses minimal graphics, while still providing variance in the visual domain. From left to right, top to bottom: CollectHealth, FourRooms, Hallway, MazeS3, OneRoom, PutNext, Sidewalk, TMaze, WallGap, YMaze. Images are upscaled to $800 \times 600$ for visual clarity.}
\label{fig:envs}
\end{figure*}

\subsection{Search}
\label{subsec:search}
Similar to \emph{Zero-shot Imitation Policy} (ZIP)~\cite{zip}, as a preliminary step we encode the expert demonstrations dataset using a pre-trained encoder $h(\cdot)$ inspired from VPT. Details of the encoder we have used are provided in Section~\ref{appdx:A}. For each trajectory $i$ and for each timestep $t$, we pass the expert state $s_{t}^{(i)}$ to obtain a latent $z_{t}^{(i)} = h(s_{t}^{(i)})$. Each latent is then paired with its corresponding action $a_{t}^{(i)}$. Thus, we define the \emph{expert latent space} $\mathcal{D}_{E}=\{(z_{t}^{(i)}, a_{t}^{(i)})\}$ as the set of encoded observation-action pairs of expert trajectories.

During inference, at each timestep $t$ we encode the current observation $o_t^{*}$, obtaining $z_{t}^{*} = h(o_t^{*})$, and retrieve the $k$-most similar latents from $\mathcal{D}_{E}$. Then, we count the number of occurrences of each action and store them in a vector $\bold{c_t}$. Finally, we model the expert action distribution as $\pi_{E}(a_{t}^{(E)}|s_t)$, where $\pi_{E}(a_{t}^{(E)} = i|s_t) = \frac{c_{t}^{(i)}}{k}$.

\subsection{Bayesian Online Adaptation}
\label{subsec:boa}
At timestep $t$, an agent (that is, a policy $\theta$) observes a state $s_t$ and selects a discrete action $a_{t}^{(\theta)}$. We can model the prior distribution $\pi_{\theta}(a_{t}^{(\theta)}|s_t)$ as a Dirichlet distribution with $K=|A|$ components and concentration vector $\boldsymbol{\alpha}_{\mathrm{prior}}$ with $\boldsymbol{\alpha}_{\mathrm{prior}, i} = \pi_{\theta}(a_{t}^{(\theta)} = i|s_t)$. That is, each pseudo-count is the probability of the corresponding action as inferred by the IL agent.

We formulate our inference problem as a Bayesian adaptation problem. Our aim is to update the beliefs of an IL agent, given a set of actions retrieved from the expert. Therefore, given the prior $\pi_{\theta}(a_{t}^{(\theta)}|s_t)$, we would need to find the likelihood $\pi_{\mathrm{E}}(a_{t}^{(E)}|a_{t}^{(\theta)}, s_t)$ to estimate the posterior. 

In our adaptation setting, $a_{t}^{(E)}$ is selected by feature similarity search between the current encoded observation $z_{t}^{*}$ and the previously encoded expert latents. Since $z_{t}^{(i)} = h(s_{t}^{(i)})$, clearly $a_{t}^{(E)}$ is conditionally dependent on $s_{t}^{(i)}$, i.e. $\pi_{\mathrm{E}}(a_{t}^{(E)}|s_t)$. Conversely, in our pipeline the action selected by the policy $a_{t}^{(\theta)}$ does not affect the choice of $a_{t}^{(E)}$. Therefore, we can safely state that, in our setting, $a_{t}^{(E)}$ is conditionally independent from $a_{t}^{(\theta)}$. Therefore
\begin{equation}
    \pi_{\mathrm{E}}(a_{t}^{(E)}|a_{t}^{(\theta)}, s_t) = \pi_{\mathrm{E}}(a_{t}^{(E)}|s_t).
\end{equation}
Thus, we can use the result of our search as likelihood. Leveraging the fact that the Dirichlet distribution is the conjugate prior of the Multinomial distribution, we can state that the posterior $\pi_{\mathrm{E}}(a_{t}^{(\theta)}|a_{t}^{(E)}, s_t)$ also follows the Dirichlet distribution with $K$ components and an updated $\boldsymbol{\alpha}_{\mathrm{posterior}} = \boldsymbol{\alpha}_{\mathrm{prior}} + \bold{c}_t$. Given that $\forall{i}, \boldsymbol{\alpha}_{\mathrm{prior}, i} \in [0, 1]$ while $\forall{i}, c_{t}^{(i)} \in \mathbb{N}$, we argue that $\boldsymbol{\alpha}_{\mathrm{posterior}}$ might be unbalanced towards $\bold{c}_t$, hence giving more importance to $\pi_{E}(a_{t}^{(E)}|s_t)$. To balance the terms, we multiply each $\boldsymbol{\alpha}_{\mathrm{prior}, i}$ by the number of searched vectors $k$, so that $\boldsymbol{\alpha}_{\mathrm{posterior}} = k \cdot \boldsymbol{\alpha}_{\mathrm{prior}} + \bold{c}_t$.

Finally, we can sample the Dirichlet posterior $\mathrm{Dir}(K, \boldsymbol{\alpha}_{\mathrm{posterior}})$ and obtain a Categorical distribution $\mathrm{Cat}(\boldsymbol{\alpha}_{\mathrm{posterior}})$. From this, the new action $\tilde{a}_{t}$ can be obtained by sampling
\begin{equation}
    \tilde{a}_{t} \sim \mathrm{Cat}(\boldsymbol{\alpha}_{\mathrm{posterior}}).
\end{equation}

\subsection{Search complexity}
\label{subsec:scompl}
Autonomous agents should act within a very short, constant amount of time. When introducing search for inference, a dependency on the complexity of the search space follows from it. In general, the more complex and dense the search space, the longer the query time.

In our work, we mitigate this dependency using \emph{faiss}~\cite{faiss}, a library for efficient search leveraging GPUs. faiss provides methods for both exact and approximate search. In our study, we encode the expert trajectories into $d$-dimensional vectors and project them in a search space with the same dimension. Then, we use (exact) L2 search to retrieve the $k$ closest samples to the current observation. In our experiments, faiss was able to handle as many as $150$ trajectories without significant delays in inference time (average search time $6.37 \pm 0.933$ms).

\section{Experiments}
\label{sec:exp}
We compare five agents, namely PPO~\cite{ppo}, BC~\cite{bc}, GAIL~\cite{gail}, ZIP~\cite{zip}, and BOA on a range of tasks from \emph{MiniWorld}~\cite{miniworld}, a modular and customizable library with 3D minimalistic graphics for fast rendering. MiniWorld features tasks of varying complexity to test diverse skills, such as navigation, memory and planning. Our experiments are run on $10$ tasks that provide an explicit terminal condition and access to a scalar reward. A more detailed description of the goals for each environment is provided in Section~\ref{appdx:B}. All environments provide $60 \times 80$ RGB observations with integer pixel values in the range $[0, 255]$. In our study, we apply minimal pre-processing to the images by simply scaling the pixel down to lie within $[0, 1]$ range. Fig.~\ref{fig:envs} shows an example of observation from each environment.

\begin{figure*}[htbp]
\centerline{\includegraphics[width=\textwidth]{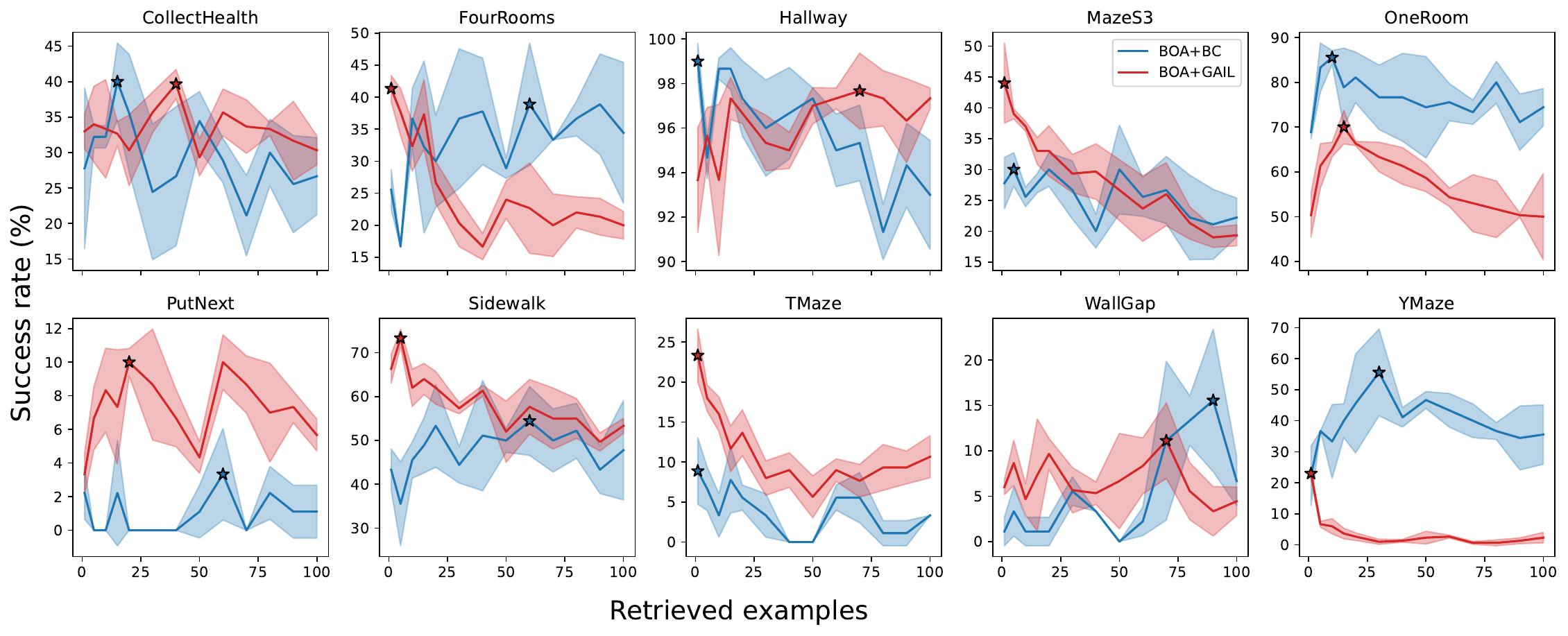}}
\caption{Mean success rate for different numbers of retrieved samples. Each graph corresponds to one environment. Measurements retrieved on $3$ runs of $30$ episodes each. The best value is marked with a star. Red lines represent a BOA agent adapting GAIL agent, while a blue line denotes a BC agent adapted with BOA.}
\label{fig:ktest}
\end{figure*}
\begin{table*}[]
\caption{Best choice of $k$ for adapted agents BOA+GAIL and BOA+BC for each environment, with associated mean success rate.}
\label{tab:ktest}
\begin{center}
\begin{tabular}{cccccccccccc}
 & & \textbf{CollectHealth} & \textbf{FourRooms} & \textbf{Hallway} & \textbf{MazeS3} & \textbf{OneRoom} & \textbf{PutNext} & \textbf{Sidewalk} & \textbf{TMaze} & \textbf{WallGap} & \textbf{YMaze} \\ \hline
                               & & & & & & & & & & & \\
\multirow{3}{*}{\textbf{\begin{tabular}[c]{@{}c@{}}BOA\\ +GAIL\end{tabular}}} & \textit{k} & $40$ & $1$ & $70$ & $1$ & $15$ & $20$ & $5$ & $1$ & $70$ & $1$ \\
                               & & & & & & & & \\
                               & \textit{\begin{tabular}[c]{@{}c@{}}Success\\ Rate (\%)\end{tabular}} & \begin{tabular}[c]{@{}c@{}}$39.67$ \\ $\pm 2.05$\end{tabular} & \begin{tabular}[c]{@{}c@{}}$41.33$ \\ $\pm 2.05$\end{tabular} & \begin{tabular}[c]{@{}c@{}}$97.67$ \\ $\pm 1.70$\end{tabular} & \begin{tabular}[c]{@{}c@{}}$44.00$ \\ $\pm 6.48$\end{tabular} & \begin{tabular}[c]{@{}c@{}}$70.00$ \\ $\pm 3.74$\end{tabular} & \begin{tabular}[c]{@{}c@{}}$10.00$ \\ $\pm 0.82$\end{tabular} & \begin{tabular}[c]{@{}c@{}}$73.33$ \\ $\pm 2.05$\end{tabular} & \begin{tabular}[c]{@{}c@{}}$23.33$ \\ $\pm 3.30$\end{tabular} & \begin{tabular}[c]{@{}c@{}}$11.11$ \\ $\pm 4.16$\end{tabular} & \begin{tabular}[c]{@{}c@{}}$23.00$ \\ $\pm 1.41$\end{tabular} \\ \hline
                               & & & & & & & & \\
\multirow{3}{*}{\textbf{\begin{tabular}[c]{@{}c@{}}BOA\\ +BC\end{tabular}}}   & \textit{k} & $15$ & $60$ & $1$ & $5$ & $10$ & $60$ & $60$ & $1$ & $90$ & $30$ \\
                               & & & & & & & & \\
                               & \textit{\begin{tabular}[c]{@{}c@{}}Success\\ Rate (\%)\end{tabular}} & \begin{tabular}[c]{@{}c@{}}$40.00$ \\ $\pm 5.44$\end{tabular} & \begin{tabular}[c]{@{}c@{}}$38.89$ \\ $\pm 9.56$\end{tabular} & \begin{tabular}[c]{@{}c@{}}$99.00$ \\ $\pm 0.82$\end{tabular} & \begin{tabular}[c]{@{}c@{}}$30.00$ \\ $\pm 2.72$\end{tabular} & \begin{tabular}[c]{@{}c@{}}$85.56$ \\ $\pm 1.57$\end{tabular} & \begin{tabular}[c]{@{}c@{}}$3.33$ \\ $\pm 2.72$\end{tabular}  & \begin{tabular}[c]{@{}c@{}}$54.44$ \\ $\pm 7.86$\end{tabular} & \begin{tabular}[c]{@{}c@{}}$8.89$ \\ $\pm 4.16$\end{tabular}  & \begin{tabular}[c]{@{}c@{}}$15.56$ \\ $\pm 7.86$\end{tabular} & \begin{tabular}[c]{@{}c@{}}$55.56$ \\ $\pm 13.97$\end{tabular} \\ \hline
\end{tabular}
\end{center}
\end{table*}

We manually collected $20$ trajectories for each task and trained all agents on the same dataset within a task. All agents used the same encoder within a task, empirically chosen to be a VPT-inspired~\cite{vpt} encoder architecture with a residual~\cite{residual} backbone and no attention obtained by training GAIL.

As a preliminary step, we study the hyperparameters of our approach, namely, the number $k$ of retrieved samples during search and the number of trajectories encoded in latent space. We test $k=\{1, 5, 10, 15, 20, 30, 40, 50, 60, 70, 80, 90, 100\}$, and estimate the best value for each task by letting a BOA agent play $3$ runs of $30$ episodes per value, using all available data. We repeat this test for both GAIL-based and BC-based BOA, as the two IL algorithms treat the action distribution differently. In particular, BC tends to be overconfident in its predictions~\cite{bc}, while GAIL predictions are usually much smoother~\cite{gail}.

To evaluate the effect of the number of encoded trajectories on performance, we manually gather an additional $130$ trajectories for the \emph{MazeS3} task, up to a total of $150$. Then, we assess the performance of BOA using $1$-$150$ trajectories, incrementing the number of encoded trajectories by $5$ each time. For this test, we use the environment-wise optimal value of $k$ found in the previous step and test each configuration over $5$ runs of $30$ episodes each.

Additionally, we compare our agents by observing their mean episodic return on each task. We enforce here that \textit{in our setting IL and adapted agents never observe the reward}, and that we collect it for the sole purpose of comparing performance. The only agent using the reward signal during training is PPO. We extensively test the agents on $6$ runs of $100$ episodes for each task.

The source code to reproduce the experiments can be found at \url{https://github.com/fmalato/online_adaptation}.

\section{Results}
\label{sec:results}
In the first part of this section we study how varying the hyperparameters of our approach affect performance. In the second subsection, we establish a numerical comparison between the tested agents.
\subsection{Hyperparameters ablation}
\label{subsec:hyperparams}
\begin{figure}[htbp]
\centerline{\includegraphics[width=\columnwidth]{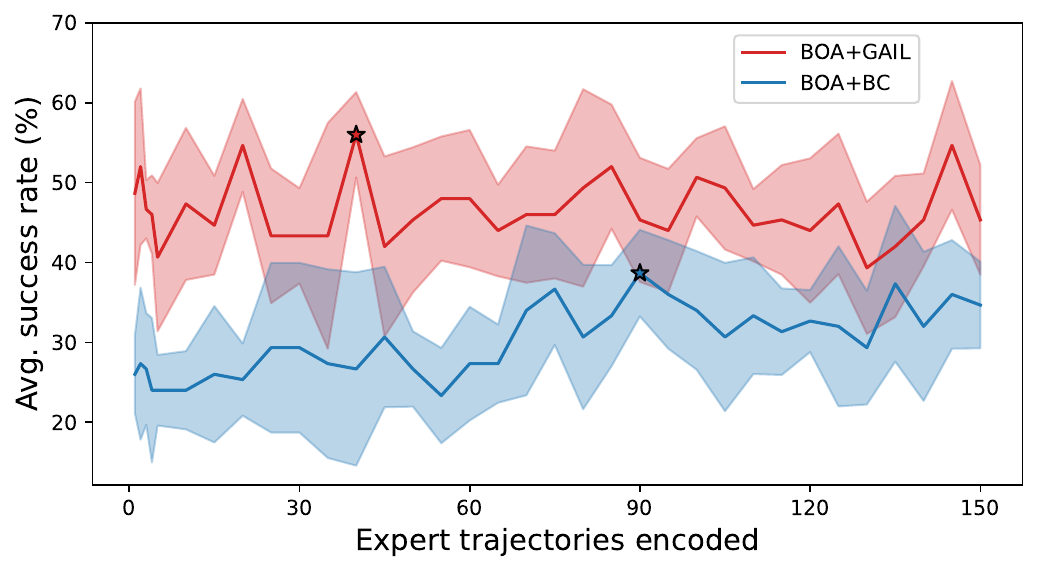}}
\caption{Mean success rate for different numbers of encoded trajectories. The test is conducted on both BOA+BC (in blue) and BOA+GAIL (red line) on $5$ runs of $30$ episodes each. $n$ varies between $1$ and $150$. We highlight the best value for each agent with a star-shaped mark.}
\label{fig:trajtest}
\end{figure}
Fig.~\ref{fig:ktest} and \ref{fig:trajtest} show the ablation study over the number of retrieved samples $k$ and the number of encoded trajectories $n$, respectively. 

\paragraph{Number of retrieved samples}
Fig.~\ref{fig:ktest} compares BOA+BC and BOA+GAIL on each environment as the number of retrieved examples $k$ increases. Overall, no significant trend is visible, and no agent consistently dominates the other. If we consider only GAIL, though, we can notice that in most environments performance is stronger when $k$ is small.

We explain this descending pattern as a joint effect of the learned latent representation and the expert trajectories. In particular, whenever the counts vector $\bold{c_t}$ has $c_{t}^{(i)} = k$ for some value of $k$, the retrieved sample is heavily polarized towards action $i$. In this case, the expert will influence $\boldsymbol{\alpha}_{\mathrm{posterior}}$ the most. On the other hand, if the expert-provided dataset includes similar observations with different actions, the effect of the expert on the posterior will, in general, be mitigated. Still in this case, if $k$ is small enough, the action distribution has a significant chance of being sharp, hence having a sensible effect over $\boldsymbol{\alpha}_{\mathrm{posterior}}$. Considering that such a descending effect is seen for the most part in maze exploration tasks, where similar observations \emph{might} carry different actions (e.g. when facing a wall, an agent can either turn left or right), the dominance for small values of $k$ is clear. On the contrary, when the environment is simple enough (e.g. Hallway) or is less likely to encounter the above-described situation, the agent show more robust performance when varying $k$.

\paragraph{Number of encoded trajectories}
In Fig.~\ref{fig:trajtest} we observe how the number of encoded trajectories affects the success rate of both adapted agents. Following from the findings of the previous paragraph, we adopt $k=1$ for BOA+GAIL and $k=5$ for BOA+BC. The dominance of BOA+GAIL over BOA+BC visible in Fig.~\ref{fig:trajtest} confirms the result of the previous experiment, where BOA+GAIL succeded $44.00\% \pm 6.48\%$ of the times, while BOA+BC yielded a mean success rate of $30.00\% \pm 2.72\%$. Overall, changing $n$ does not seem to affect any of the adapted agents, even though a faint trend seems to suggest that BOA+BC would benefit from additional data.

The best performance of BOA+GAIL is $56.00\% \pm 5.96\%$ ($n=40$), while the worst mean success rate is $39.34\% \pm 9.25\%$ ($n=130$). As for BOA+BC, the best and worst results are $38.67\% \pm 6.06\%$ ($n=90$) and $24.66\% \pm 3.80\%$ ($n=50$) respectively.

\subsection{Numerical performance}
\label{subsec:numres}
\begin{figure*}[htbp]
\centerline{\includegraphics[width=\textwidth]{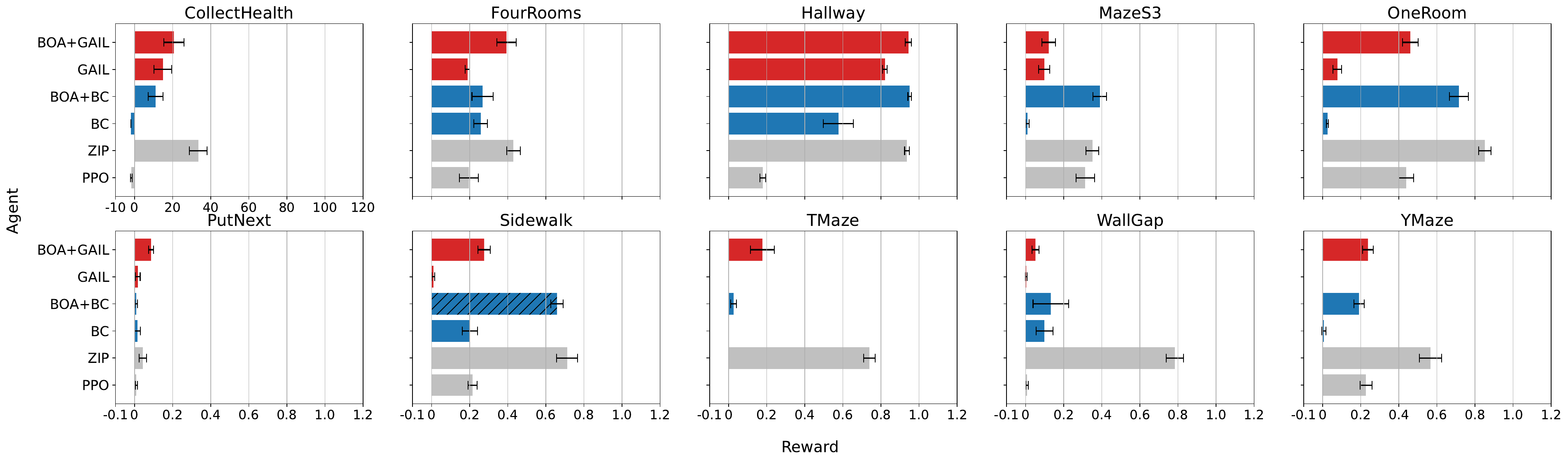}}
\caption{Average reward comparison over the $10$ selected tasks. Higher is better. In all environments, the reward is in the range $[0, 1]$ except for \emph{CollectHealth}, where the reward is in $[-2, +\infty)$. We highlight BOA agents with striped bars and link them to the corresponding IL agent by matching the color of the bar. Grey bars represent baseline methods.}
\label{fig:reward}
\end{figure*}
Fig.~\ref{fig:reward} shows the average return for each tested agent. At first glance, BOA agents either outperform or match the performance of their corresponding IL agent in all environments. Notably, the adaptation generally improves the reward even when the underlying IL policy catastrophically fails.

From Fig.~\ref{fig:reward} we see how PPO is able to complete a good portion of the exploration tasks, except for TMaze and WallGap. Still, PPO does not achieve reliable performance in any of them. We hypothesize this is due to the sparse nature of the reward. Our other baseline, ZIP, achieves strong performance on every task except PutNext. We believe this follows from the inherent difficulty of the environment that, unlike all the others, requires two sequential steps to achieve the goal.

GAIL and BC solve some of the tasks with acceptable results and fail in others. We suppose that in those cases, BC fails in environments where data does not extensively encapsulate the complexity of the environment, while GAIL failures are due to its inherent instability.

As expected, both BOA agents improved the capabilities of their corresponding IL policies. BOA+BC achieves almost-zero performance only in TMaze and PutNext. We suspect this happens for two reasons: first, BC completely fails on the task; second, the IL policy assigns a very high probability to the selected action, leading the adaptation to fail as well. When this is not true, such as for GAIL, we see how the adaptation still improves the results.

These findings confirm our hypothesis: while selecting actions directly from the expert distribution is likely to yield better results, adapting IL agents with search on the expert data improves performance. Nonetheless, while ZIP blindly copies actions based on past situations, BOA always takes into account the possibility of different situations. In a more unpredictable and complex scenario we expect BOA to perform better than ZIP, as suggested by the \textit{PutNext} task. We motivate this claim with the following reasoning.

ZIP assumes that an expert can provide optimal solutions to any situation within a specified task. Therefore, ZIP will reliably solve tasks that are represented at least once within the expert dataset. As the complexity of the task increases, ZIP will either require a much larger expert dataset or encounter limitations. On the contrary, an IL policy should be able to generalize its knowledge, thus being capable of addressing some shortcomings of the expert dataset. In such setting, problems are likely to arise whenever a certain situation is under-represented in the dataset.

BOA complements the two approaches: on one hand, it is capable of addressing the under-representation problem of IL methods thanks to search; on the other hand, it allows generalization in situations where the encoded experience is contradictory. Still, BOA needs a softer search mechanism than ZIP to allow generalization. As such, BOA might fall short of ZIP in simple tasks, but is bound to improve on complex ones such as \textit{PutNext}. Most importantly, BOA \textit{always} improve its underlying policy thanks to search, while keeping generalization capabilities intact.

\subsection{Perceptual evaluation}
\label{subsec:perceptual}
As an additional set of experiments, we evaluate the adapted agents perceptually. To support the statements of this subsection, we have released a video on our YouTube channel (visible at \url{https://youtu.be/WoWalj4CVmM}). Moreover, to keep the amount of visual material limited and the evaluation concise, we evaluate three environments with diverse purposes, namely FourRooms for navigation, PutNext for the sequential tasks, and CollectHealth to evaluate the survival skills. For each environment, we watch $10$ episodes played by ZIP, followed by BC and GAIL, and finally by BOA+BC and BOA+GAIL.

\paragraph{FourRooms}
According to our evaluation, ZIP possesses the best "human-like" skills: if it gets stuck in front of a wall, it will most likely recover and can confidently pass through doors. Still, when seeing the red box, in a significant number of instances, ZIP has ignored it. This is a consequence of action copy: if a new search is triggered while the red box is in sight, ZIP will reach the goal. Otherwise, the new search will likely lead to more exploration.

Despite being tested as stochastic, BC will likely get stuck. The agent possesses decent navigation skills, but the overconfidence in its prediction will make it often fixate on a single action. As expected, GAIL possesses better navigation skills and can confidently explore. While it also gets stuck at times, it is much more likely to recover than BC.

The first clear difference between both BOA agents and the other agents is that whenever the red box is spotted, they actively aim for it. Judging from the behavior of the plain IL policies, this is likely an effect of the adaptation. Between BOA+BC and BOA+GAIL, the latter possesses the best navigation skills. While BOA+BC can confidently navigate (resembling ZIP sometimes), it can not overcome door obstacles.

\paragraph{PutNext}
ZIP can confidently explore the room. Additionally, it usually picks up some block and drops it next to another. Still, the case where those two coincide with a yellow and red block is low, hence probability of success is even lower.

BC aims for a block most of the time but then fails to pick it up. We suspect this is a consequence of the unbalance in data, as pick up and drop actions are well under-represented with respect to movement actions. GAIL also seems to be aiming for blocks and can sometimes pick them up, but its actions resemble more a slightly conditioned random policy than a confident one. We suspect that in this case, the training collapsed.

BOA agents show interesting and complementary behavior. Specifically, while this would need to be confirmed further with additional experiments, BOA+BC seems to actively aim for the yellow box and, sometimes, picks it up. Then, it is unable unlikely to drop it. On the contrary, BOA+GAIL can pick up any block and, sometimes, drop it near another. Still, it does not seem to actively look for the yellow one.

\paragraph{CollectHealth}
ZIP shows confident exploratory behavior and is capable of picking up a couple of health kits before dying.

BC mostly turns right. At times, it attempts to aim for a health kit, but the rapidity of the environment does not allow it to pursue the goal. GAIL is more confident in walking towards health kits, but only sometimes manages to pick one up.

BOA+BC corrects the turning behavior of plain BC and allows the agent to reach, sometimes, for a health kit. Similarly, BOA+GAIL can aim for health kits, but only sometimes it can reach one.

\section{Conclusions}
\label{sec:concl}
We have presented Bayesian online adaptation (BOA), a hybrid search-learning approach for adapting an imitation learning agent in real-time via search on expert-provided data. We have shown how our approach performs better than pure IL agents while falling short of action copy. Regardless, BOA agents carry the advantage of potentially being able to adapt to unpredictable situations, leveraging their underlying IL policy. Future research could investigate the effect of adaptation on RL agents such as PPO, or how to improve training time of RL agents by using adapted actions during training. Another direction could study balancing the effect of the adaptation based on the relevance of each retrieved sample. Finally, combining BOA with classical RL algorithms could prove adaptation as a versatile strategy to boost learning and improve sample efficiency.

Other than improving performance, BOA implicitly addresses the problem of explainability in machine learning. While being far from solving it, such a hybrid approach can always track the effect of the expert-retrieved samples, hence providing additional insight into the action selection process. For instance, during our perceptual evaluation, we noticed that in some instances the effect of the adaptation was immediately clear. Thus, an interesting future research direction could leverage this intuition to enhance explainable agents.

\section{Appendix A - Encoder details}
\label{appdx:A}
As stated in Section~\ref{subsec:search}, in our study we use a VPT-inspired encoder $h(\cdot)$ to build the latent search space. We have designed our encoder with two purposes in mind: on one hand, we want a temporally extended representation to better characterize each state; on the other hand, we want to keep the architecture as shallow as possible, to keep our architecture easily trainable on consumer hardware with small datasets.

The original VPT architecture is built on top of the Inverse Dynamics Model (IDM)~\cite{vpt} for predicting actions from observations. The input to both models is a stack of $128$, $128\times128$ RGB frames. IDM features a 3D convolution followed by three residual stacks and two fully connected layers with $256$ and $4096$ activations, respectively. Each residual stack is composed of a 2D convolution with max pooling and two residual blocks as described in~\cite{residual}. The output is then forwarded to four transformer heads which build long-term relationships between frames. With respect to IDM, VPT removes the 3D convolution layer.

In our work, we use the same architecture as IDM, up to the two fully connected layers. Differently from IDM, our FC layers feature $2048$ and $1024$ activations. Additionally, we retain the 3D convolution layer but use a stack of $4$, $60\times80$ RGB images as input to encapsulate temporally close dependencies, similarly to~\cite{atari}. As a result, our encoder is composed of ~$50M$ parameters, a $90\%$ reduction with respect to IDM/VPT.
\section{Appendix B - Environments description}
\label{appdx:B}
We provide a short description for each of the $10$ tasks used in the study. An example of each environment is provided in Fig.~\ref{fig:envs}.

We use $7$ unique navigation tasks in our experiments. They all share the same goal, that is, reaching a randomly-spawned red box. The reward is $+(1 - 0.2 * \frac{t}{T})$ upon reaching the red box, where $t$ is the current number of timesteps and $T$ is the maximum duration of an episode; $0$ otherwise. The action space is discrete, 3-dimensional featuring actions \{\textit{left}, \textit{right}, \textit{forward}\}. We provide a brief description of each.
\begin{itemize}
    \item \textbf{FourRooms:} Four rooms with chess floor tiles, grey walls and ceiling, connected in a quadrant pattern.
    \item \textbf{Hallway:} One rectangular room with chess floor tiles, grey walls and ceiling.
    \item \textbf{MazeS3:} Three randomly-generated rooms connected with corridors. MazeS3 features chess tiles, red brick walls and grey ceiling.
    \item \textbf{OneRoom:} One big squared room with chess floor tiles, grey walls and ceiling.
    \item \textbf{TMaze:} Two corridors connected in a T pattern. Goal position is randomly spawned on either branch of the T.
    \item \textbf{WallGap:} Two rooms with no ceiling connected with a gap in a wall. Agent spawns in one room, red box in the other.
    \item \textbf{YMaze:} Similar to TMaze, but arranged in the shape of a "Y".
\end{itemize}
Additionally, we use three tasks with different rules. Their descriptions are as follows.\\
\textbf{Name:} \textit{CollectHealth}
\begin{itemize}
    \item \textbf{Description:} The environment is composed of a single room with green floor and grey walls and ceiling. Scattered all over the floor are health kits.
    \item \textbf{Goal:} Survive as long as possible.
    \item \textbf{Action space:} Discrete, 8-dimensional. Actions: \{\textit{left}, \textit{right}, \textit{forward}, \textit{backwards}, \textit{pick up}, \textit{drop}, \textit{toggle}, \textit{complete}\}.
    \item \textbf{Rules:} Agent has an initial health of $100$. After each timestep, health decreases by $2$. If health reaches $0$, the agent dies. Whenever an agent collects a health kit, its health is restored back to $100$. Collecting a health kit requires actively performing a \textit{pick up} action.
    \item \textbf{Reward}: $+2$ for each timestep; $-100$ for dying.
\end{itemize}
\textbf{Name:} \textit{PutNext}
\begin{itemize}
    \item \textbf{Description:} One big, squared room with chess tiles, grey wall and ceiling, filled with colored boxes. 
    \item \textbf{Goal:} Place the yellow box next to the red box.
    \item \textbf{Action space:} Discrete, 8-dimensional. Actions: \{\textit{left}, \textit{right}, \textit{forward}, \textit{backwards}, \textit{pick up}, \textit{drop}, \textit{toggle}, \textit{complete}\}.
    \item \textbf{Rules:} An agent reaches the yellow box and performs a \textit{pick up} action. Then, the agent must navigate to the red box and perform a \textit{drop} action.
    \item \textbf{Reward}:  $+(1 - 0.2 * \frac{t}{T})$ when red and yellow boxes are close; $0$ otherwise.
\end{itemize}
\textbf{Name:} \textit{Sidewalk}
\begin{itemize}
    \item \textbf{Description:} One rectangular room opened on one side. The opened side is delimited with traffic cones. Red-brick walls, grey floor and no ceiling.
    \item \textbf{Goal:} Reach the red cube within the time limit.
    \item \textbf{Action space:} Discrete, 3-dimensional. Actions:  \{\textit{left}, \textit{right}, \textit{forward}\}.
    \item \textbf{Rules:} Navigate the maze to reach the goal location. If the agent steps out of the sidewalk, it dies.
    \item \textbf{Reward}: $+(1 - 0.2 * \frac{t}{T})$ when the goal is reached; $0$ otherwise.
\end{itemize}
\end{document}